\documentclass{article}

\usepackage{microtype}
\usepackage{graphicx}
\usepackage{subfigure}
\usepackage{booktabs} 

\usepackage{hyperref}

\usepackage[accepted]{icml2025}

\usepackage{amsmath}
\usepackage{amssymb}
\usepackage{mathtools}
\usepackage{amsthm}

\usepackage[capitalize,noabbrev]{cleveref}

\theoremstyle{plain}

\theoremstyle{definition}

\theoremstyle{remark}

\usepackage[disable,textsize=tiny]{todonotes}

\icmltitlerunning{Lessons from Developing and Maintaining Open-Source AI Evaluations}

\begin{document}

\twocolumn[

\icmltitle{Developing and Maintaining an Open-Source Repository \\
           of AI Evaluations: Challenges and Insights}



\icmlsetsymbol{equal}{*}

\begin{icmlauthorlist}
\icmlauthor{Alexandra Abbas}{comp}
\icmlauthor{Celia Waggoner}{comp}
\icmlauthor{Justin Olive}{comp}
\end{icmlauthorlist}

\icmlaffiliation{comp}{Arcadia Impact, London, United Kingdom}

\icmlcorrespondingauthor{Justin Olive}{justin@arcadiaimpact.org}

\icmlkeywords{Open Source, AI Evaluation}

\vskip 0.3in
]



\printAffiliationsAndNotice{}  

\begin{abstract}
AI evaluations have become critical tools for assessing large language model capabilities and safety. This paper presents practical insights from eight months of maintaining $inspect\_evals$, an open-source repository of 70+ community-contributed AI evaluations. We identify key challenges in implementing and maintaining AI evaluations and develop solutions including: (1) a structured cohort management framework for scaling community contributions, (2) statistical methodologies for optimal resampling and cross-model comparison with uncertainty quantification, and (3) systematic quality control processes for reproducibility. Our analysis reveals that AI evaluation requires specialized infrastructure, statistical rigor, and community coordination beyond traditional software development practices.
\end{abstract}

\section{Introduction}
\label{introduction}

AI evaluations have emerged as a valuable way to assess the capabilities and safety of large language models (LLMs). The UK AI Security Institute (UK AISI) identifies AI evaluations as a critical tool to advance our understanding of model capabilities and prevent catastrophic AI risks \yrcite{UKAISI2024research}, recognizing their fundamental importance in developing safe and beneficial AI systems.

This paper presents learnings and challenges from the maintenance of the $inspect\_evals$ repository \cite{inspectevalsrepo} since its publication 8 months ago. $inspect\_evals$ is a repository of community-contributed LLM evaluations for $inspect\_ai$ \cite{inspectairepo}, created in collaboration by the UK AISI, Arcadia Impact, and the Vector Institute. The repository houses a diverse range of (70+) evaluations, including assessments of scientific knowledge such as GPQA \cite{rein2023gpqa}, offensive cyber capabilities such as Cybench \cite{zhang2024cybench}, and the robustness of safeguards such as AgentHarm \cite{andriushchenko2024agentharm}.

AI evaluations have long presented challenges that traditional software engineering practices struggle to address, and these challenges are intensifying as evaluations become more numerous, complex, and consequential. Through our experience with $inspect\_evals$, we have encountered technical, methodological, and community-driven obstacles that can provide valuable insights for the broader open-source AI evaluation community. This paper reviews these challenges, proposes solutions to some, and identifies others as open questions requiring further research and community coordination.

\section{Repository Overview}
\label{repository_overview}

$inspect\_evals$ is a collaborative space where the evaluation community can contribute high-quality standardized assessments, breaking down traditional silos across organizations and facilitating wider adoption of rigorous evaluations.

The repository serves three primary audiences: researchers working on scaling laws and evaluation science, analysts processing model performance data for decision-makers, and technical teams running internal evaluations against industry benchmarks.

$inspect\_evals$ organizes evaluations across nine categories: Agents, Assistants, Coding, Cybersecurity, Knowledge, Mathematics, Multimodal, Reasoning, and Safeguards, providing comprehensive coverage of modern AI capabilities and safety considerations.

\section{Learnings and Challenges}

Maintaining an open-source repository of AI evaluations introduces new challenges beyond traditional software development. Key challenges include:

\begin{itemize}
\item The rapid pace of new evaluation publications combined with the absence of centralized evaluation sources complicates discovery and curation efforts;
\item Assessing the quality, credibility, and usefulness of evaluations is resource intensive, requiring careful analysis, and in many cases, domain expertise;
\item Incomplete or low-quality documentation of the evaluation methodology (including the reference implementation) creates challenges in reproducing the results.
\end{itemize}

\subsection{Evaluation Selection and Prioritization}

With evaluations being published continuously, it is necessary to capture and process this constant stream of new releases. To this end, we developed a roadmap that prioritizes evaluations, assigns them an implementation difficulty score, and identifies which capabilities they cover. One factor that influences our prioritization is filling gaps in the least-covered areas. Sources of evaluations include technical reports, The AI Evaluation Substack \cite{aievaluationsubstack}, and suggestions from the community. Some of the criteria that we consider when sourcing and prioritizing evaluations include the following.

\begin{itemize}
\item Well-established with research citations
\item Credibly sourced from major labs or academic groups
\item Challenging for frontier models with distinguishable performance
\item Agentic/task-based over simple Q\&A (question-and-answer)
\item Clearly scoped with verifiable methodologies
\item Comparable with existing baseline results
\end{itemize}

\subsection{Contributor Management}

To help keep up with the rapid pace of new evaluation publications and to help upskill the open-source community, we periodically facilitate cohorts of volunteers to implement benchmarks in $inspect\_evals$. We organize volunteers into 5-week cohorts with one Technical Project Manager (TPM) overseeing 5-10 engineers. Implementation involves porting evaluations from papers to $inspect\_evals$.

\begin{figure}[ht]
\vskip 0.2in
\begin{center}
\centerline{\includegraphics[width=\columnwidth]{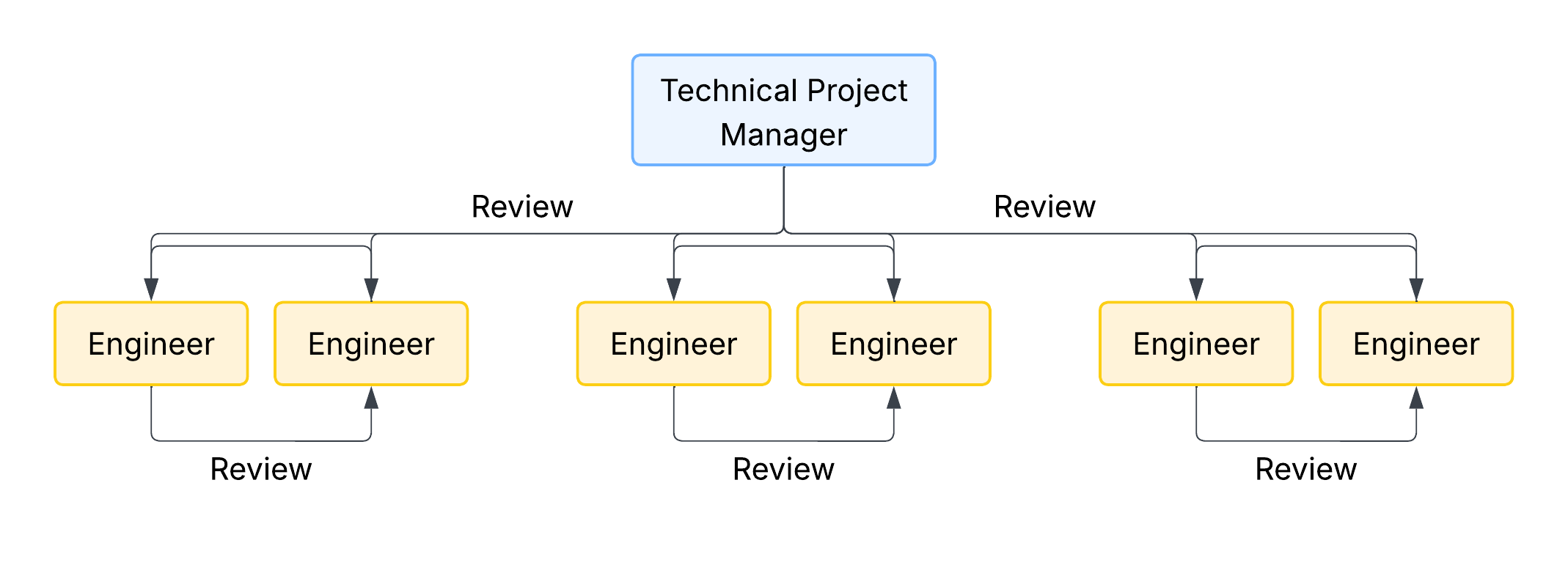}}
\caption{Cohort management structure showing one Technical Project Manager overseeing multiple Software Engineers with bidirectional review processes for coordinated evaluation implementation and scalability.}
\label{contributor-chart}
\end{center}
\vskip -0.2in
\end{figure}

Our program follows structured phases: an onboarding week, a design phase where contributors produce a Benchmark Development Plan that is reviewed by the TPM and a peer, the implementation phase for creating draft Pull Requests (PRs) with TPM and peer review, and the final PR submission with evaluation reports.  This structure addresses the educational gap and enables scalable coordination of volunteer contributions through agile project management including standups, Kanban boards, and detailed documentation.

Outside of the benchmarks cohort program, we create GitHub issues to signal to the open-source community which evaluations we would be particularly excited to see implemented in $inspect\_evals$. We provide support to individual contributors who pick up these issues by answering questions and reviewing their code.

\subsection{Quality Assurance}

Ensuring reliable AI evaluations is a challenge that requires systematic quality control to address both the correctness of implementation and the methodological validity.

\textbf{Contributor guidelines and review process.} Participants in the volunteer cohorts are provided with a Software Requirement Specification that outlines a 5-week delivery timeline with weekly deliverables. Contributors must first produce a Benchmark Development Plan to be reviewed by their TPM before implementation. The Benchmark Development Plan details the evaluation scope, methodology, scoring mechanisms, and verification strategy. This upfront design prevents implementation drift and ensures alignment with evaluation objectives.

For individual contributors outside of the volunteer cohorts we developed a Contribution Guide which established standardized procedures for submitting new evaluations.

\textbf{Verification and testing methods.} Contributors participating in our volunteer cohorts work in pairs to verify each other's implementations on data subsets, with TPM oversight. For all contributions to $inspect\_evals$, automated unit and integration tests are run in CI pipelines to validate core functionality and framework compliance, and logs are manually reviewed to verify that frontier models or humans can complete evaluation samples. This prevents tool failures that artificially degrade model performance and identifies potential reward hacking behaviors.

\subsubsection{Testing AI Evaluations}

\textbf{Standards and requirements.} We follow the UK AISI Autonomous Systems Evaluation Standard \cite{UKAISI2024autonomousevalstandard}, which requires unit tests for custom tools and scorers, plus verification tests for dynamic resource access like Hugging Face datasets. We also conduct manual peer testing to ensure that the benchmarks successfully execute end-to-end, though there are plans to migrate this to automated CI processes.

\textbf{Result validation.} Frontier models are run on evaluation subsets to verify that results match reference implementations within +/-5\% deviation. Deviations require investigation, which is fairly common due to authors not publishing exact model configuration details, like temperature or specific endpoints. Manual tests require re-execution when implementations update, creating substantial overhead we seek to automate.

\subsection{Reproducibility vs. Cost: Finding Balance}

Validating evaluation implementations against reference results requires careful consideration of computational costs and statistical significance. Contributors are instructed to run evaluations on the entire dataset when feasible, but often resort to random subsets (maximum \$100 or 10-20\% of data) due to cost constraints. This approach introduces noise and risks missing rare but important cases that could affect the validity of the implementation.

Agentic evaluations present unique challenges due to small sample sizes with heterogeneous tasks, where agents may fail to complete any samples. In such cases, contributors are required to manually examine logs to make sure that agents aren't failing on obvious fixable issues. 

Q\&A evaluations may be run multiple times to establish statistical significance, allowing us to calculate confidence intervals and test whether performance differences exceed random variation. However, this approach must also balance cost considerations with reliability requirements. See Section~\ref{statistical-methodology} for more details.

\subsection{Statistical Methodology}
\label{statistical-methodology}

\todo{Review inline equations for correctness}

Language models exhibit non-deterministic behavior, making single evaluations potentially unreliable since repeated runs on the same sample may yield different results. This raises the critical question of how many evaluation runs are necessary to obtain statistically reliable results while managing computational costs effectively.

\textbf{Resampling methodology.} Models can be resampled on identical samples using the epoch parameter in $inspect\_ai$ to analyze result variability. To determine optimal epoch counts for statistical significance, a methodology is used that involves sampling 100 random samples and evaluating models across 10 epochs \cite{resamplinganalysis2025}. From these measurements, two key statistical values are computed: $E[\sigma_i^2]$ (the mean of individual sample variances) and $\text{Var}(\bar{x})$ (the variance of sample means).

\textbf{Optimal resampling determination.} Using Miller's equation $E[\sigma_i^2]/K \ll \text{Var}(\bar{x})$, we can determine $K$, the optimal number of resampling iterations \cite{miller2024statistical}. Once this equation is satisfied, additional resampling provides diminishing returns, indicating we have identified the computational sweet spot where reliable results are achieved without excessive resource expenditure. This approach provides a principled framework for balancing statistical rigor with practical cost constraints in AI evaluation workflows.

\subsection{Cross-Model Comparison}

AI evaluations play a critical role in comparing model capabilities, but establishing reliable and statistically robust comparisons presents methodological challenges. The fundamental question becomes: How can we ensure that cross-model comparisons yield meaningful, comparable results that accurately reflect relative performance differences?

\subsubsection{Elicitation Methods and Fair Comparison}

Different elicitation techniques such as prompt engineering and agent scaffolding can significantly impact model performance, with optimal approaches varying across models. The space of possible elicitation techniques is vast and computationally infeasible to explore exhaustively. While unelicited results might appear most fair for comparison, they may inadvertently benefit certain models over others due to inherent prompt biases. We opted for using unelicited model responses to maintain consistency, though this choice involves trade-offs between fairness and maximizing individual model potential.

\subsubsection{Statistical Significance Testing Between Model Scores}

Two complementary statistical approaches are implemented for model comparison, each providing different insights depending on assumptions about response correlation \cite{miller2024statistical, pairwiseanalysis2025}.

\textbf{Unpaired Analysis.} This method treats model responses as independent variables, computing score differences $\Delta S = S_A - S_B$ with combined standard error $SE_{unpaired} = \sqrt{SE_A^2 + SE_B^2}$. The 95\% confidence interval becomes $\text{CI} = [\Delta S - 1.96 \times SE_{unpaired}, \Delta S + 1.96 \times SE_{unpaired}]$. While conceptually straightforward, this approach yields wider confidence intervals and reduced statistical power by ignoring potential correlations between model responses.

\textbf{Paired Analysis.} This more sophisticated approach leverages correlation between model responses on identical inputs by analyzing per-question score differences $d_i$. For each question where both models were evaluated, we compute the difference in scores. The mean score difference is calculated as $\bar{d} = \frac{1}{N} \sum_{i=1}^{N} d_i$, where $N$ is the number of questions. The standard error becomes $SE_{paired} = \frac{SD_d}{\sqrt{N}}$, where $SD_d$ represents the standard deviation of score differences. Using this standard error, we calculate the 95\% confidence interval as $CI = [\bar{d} - 1.96 \times SE_{paired}, \bar{d} + 1.96 \times SE_{paired}]$. This method typically produces narrower confidence intervals and greater statistical precision by removing variability due to question-specific difficulty differences.

\textbf{Limitations of multi-model comparison.} Robust simultaneous comparison of more than two models remains statistically challenging due to the exponential growth of pairwise relationships.

Leaderboards that simply compare mean scores across multiple models without reporting standard errors or confidence intervals can yield misleading results because they treat all observed score differences as meaningful, regardless of statistical significance. Without uncertainty quantification, small performance gaps that fall within measurement noise are presented as definitive rankings, potentially leading to incorrect conclusions about relative model capabilities.

\section{Future Directions}

The challenges identified in maintaining $inspect\_evals$ point toward several promising research directions and infrastructure improvements that could benefit the broader AI evaluation community.

\textbf{Automated validation of evaluation implementations.} Current CI pipelines in $inspect\_evals$ only run basic tests, while manual testing handles integration testing and ensures agentic evaluations do not fail on obvious mistakes like tool bugs. Future work should automate these validation processes to reduce manual overhead and eliminate the need to repeat manual tests whenever evaluations are updated.

\textbf{Private test sets with a centralized trusted protocol.} Evaluation contamination through training data leakage remains a persistent challenge, and there is currently no standardized way of accessing private test sets without contamination risk. Researchers use different access restriction methods, making it difficult for $inspect\_evals$ to obtain results on these test sets. Developing a centralized, trusted protocol for maintaining and accessing private test sets could preserve evaluation integrity while enabling meaningful model comparisons.

\textbf{Duplication mitigation through a trusted evaluation log database.} Organizations currently waste computational resources by independently reproducing evaluation results. A collaborative database of trusted evaluation results with proper provenance tracking could reduce redundant computation while improving result reliability.

\textbf{Multi-dimensional capability mapping beyond categorical organization.} The current nine-category structure in $inspect\_evals$ may be too rigid to capture nuanced AI capabilities. Dynamic capability mapping using embedding spaces and graph-based representations could better identify emergent abilities and capability interactions \cite{zhou2025generalscales}.

\textbf{Cost-efficient evaluation sampling strategies.} When validating evaluation implementations added to $inspect\_evals$, current random sampling with arbitrary cost caps introduces noise and risks missing important cases without running on entire expensive datasets. Advanced sampling methodologies using stratified sampling based on difficulty levels or task types could optimize validation coverage while minimizing computational expense for implementation verification.

\section{Conclusion}

This paper documents the practical challenges and methodological insights gained from maintaining $inspect\_evals$ over eight months of active development.

The rapid pace of new evaluations, incomplete paper details, and buggy reference implementations create systematic barriers that individual organizations cannot solve alone. We found that structured community management through cohort programs effectively scales contributions while maintaining quality. Statistical analysis reveals that common practices like single evaluation runs and mean-only leaderboard comparisons are methodologically unsound. Paired statistical analysis and proper uncertainty quantification are essential for reliable model comparisons.

Implications for practitioners and the field highlight several critical needs. Practitioners should not trust single evaluation runs or mean-only comparisons and should budget for proper statistical validation alongside implementation costs. The broader field requires dedicated evaluation infrastructure and methodologies in addition to adapted software practices, with community coordination being essential for addressing shared challenges such as evaluation contamination and cost-efficient validation.

The challenges identified require coordinated effort across academic institutions, AI safety organizations, and industry partners. We call upon the community to collaborate on developing a shared infrastructure. Investment in evaluation infrastructure is critical for AI safety, and standardization efforts require sustained, coordinated funding to succeed.

\section*{Impact Statement}

This paper presents work whose goal is to advance the field of AI evaluations and safety. Our work on maintaining and improving AI evaluations has several potential societal consequences worth highlighting.

\textbf{Positive impacts.} By improving the quality, accessibility, and reliability of AI evaluations, this work contributes to better understanding of AI capabilities and limitations, which is essential for safe AI development and deployment. Standardized evaluation practices can help identify potential risks before they manifest in deployed systems, and our statistical methodologies can reduce misleading comparisons that might inform poor decision-making about AI system capabilities.

\textbf{Potential risks.} Centralized evaluation repositories could become single points of failure or create evaluation monocultures that miss important capability dimensions. Additionally, if evaluation implementations contain biases or errors, widespread adoption through repositories like $inspect\_evals$ could propagate these issues across the research community. The push for standardization might also inadvertently stifle innovation in evaluation methodologies.

\textbf{Dual-use considerations.} While our work aims to improve AI safety through better evaluation, the same infrastructure and methodologies could potentially be used to optimize AI systems for harmful capabilities. We mitigate this risk by focusing on safety-relevant evaluations and working closely with AI safety organizations, but acknowledge that evaluation tools are inherently dual-use.

\section*{Acknowledgements}

We thank JJ Allaire (UK AISI) and Charles Teague (RAND) for their mentorship and trust throughout the course of this project. We are grateful to Evan Miller (Anthropic) for his guidance on cross-model statistical testing, and to Jean-Stanislas Denain (Epoch AI) for his feedback on the $inspect\_evals$ dashboard. We also thank Matt Fisher, Alexander Putilin, Shaheen Ahmed-Chowdhury, and Nelson Gardner-Challis for their invaluable support as maintainers and Technical Project Managers, and Joe Hardie for overseeing operations.

We are deeply appreciative of all open-source contributors whose efforts made this project possible. Development and maintenance were funded by the UK AISI, and the coordination of open-source volunteer cohorts was supported by Open Philanthropy.


\bibliography{paper}
\bibliographystyle{icml2025}

\newpage
\appendix
\onecolumn

\section{Evaluation Cost Estimates}

Table~\ref{eval-costs} shows the estimated costs for running various evaluations across different models. These estimates highlight the significant cost differences between models, with GPT-4.5 being substantially more expensive than open-source alternatives like Llama-3.2-90B or commercial alternatives like DeepSeek-R1. The cost variations are particularly pronounced for more complex evaluations like GAIA Level 3, where GPT-4.5 costs over 80 times more than DeepSeek-R1.

\begin{table}[H]
\caption{Estimated costs (USD) for running evaluations on different models.}
\label{eval-costs}
\vskip 0.15in
\begin{center}
\begin{small}
\begin{sc}
\begin{tabular}{lccc}
\toprule
Evaluation & GPT-4.5 & DeepSeek-R1 & Llama-3.2-90B \\
\midrule
GPQA-Diamond \cite{rein2023gpqa} & \$81.27 & \$1.09 & \$0.57 \\
Cybench \cite{zhang2024cybench} & \$197.14 & \$1.57 & \$2.26 \\
GAIA Level 3 \cite{mialon2023gaia} & \$1,265.02 & \$9.37 & \$15.11 \\
\bottomrule
\end{tabular}
\end{sc}
\end{small}
\end{center}
\vskip 0.1in
\begin{small}
\textit{Note:} Estimates are based on input/output token counts and reflect pricing as of April 23, 2025, for endpoints: openai/gpt-4.5-preview-2025-02-27, openrouter/deepseek/deepseek-r1, and openrouter/meta-llama/llama-3.2-90b-vision-instruct.
\end{small}
\vskip -0.1in
\end{table}


\end{document}